\documentclass[conference]{IEEEtran}

\usepackage{cite}

\ifCLASSINFOpdf
\else
\fi
\usepackage{amsmath}
\usepackage{array}

\ifCLASSOPTIONcompsoc
 \usepackage[caption=false,font=normalsize,labelfont=sf,textfont=sf]{subfig}
\else
 \usepackage[caption=false,font=footnotesize]{subfig}
\fi
\usepackage{multirow}
\usepackage{graphicx}

\begin{document}
%
\title{Measurement-wise Occlusion in Multi-object Tracking}

\author{Michael Motro and Joydeep Ghosh \\%
\IEEEauthorblockA{Department of Electrical and Computer Engineering\\
University of Texas at Austin, USA}
}


\maketitle

\begin{abstract}
Handling object interaction is a fundamental challenge in practical multi-object tracking, even for simple interactive effects such as one object temporarily occluding another. We formalize the problem of occlusion in tracking with two different abstractions. In object-wise occlusion, objects that are occluded by other objects do not generate measurements. In measurement-wise occlusion, a previously unstudied approach, all objects may generate measurements but some measurements may be occluded by others. While the relative validity of each abstraction depends on the situation and sensor, measurement-wise occlusion fits into probabilistic multi-object tracking algorithms with much looser assumptions on object interaction. Its value is demonstrated by showing that it naturally derives a popular approximation for lidar tracking, and by an example of visual tracking in image space.
\end{abstract}


%
\IEEEpeerreviewmaketitle

\section{Introduction}
Recent applications of robotics, such as intelligent consumer vehicles, require an understanding of their surroundings on par with a human's. This is currently achieved via maximizing information intake at all times, combining high-resolution sensors like multi-laser rotational lidars with powerful computers and substantial context such as 3D maps. Lower-resolution sensors and weaker computation could perhaps achieve the necessary level of understanding at a lower cost, but require a system that accurately and completely handles any uncertainties. The framework of multi-object tracking achieves this by modeling the environment as a set of objects whose presence, location, and characteristics follow potentially interdependent probability distributions. A carefully designed model can intrinsically perform complex tasks such as combining information from different points of view, correctly reasoning about yet-undetected objects, and quantifying uncertainty in its predictions.

Not every property of real multi-object systems can be easily formulated in this framework. For example, the majority of models treat the motion of each object as independently distributed, though many tracking applications feature objects that dynamically interact, for instance by following each other. Similarly, these models do not always enforce inter-object constraints such as that two objects cannot occupy the same space, though there are some ways to implement such constraints \cite{grid_detailed}. Multi-object tracking models also typically assume that sensory information is the accumulation of individual information from each object within the sensor's view. In practice, measurements may be a more complex result of several nearby objects. The clearest example of this is termed occlusion: sensors relying on line-of-sight will not receive information from objects that are behind other objects.

Occlusion is a simple concept but has no standard treatment for multi-object trackers. Offline visual tracking techniques often treat occlusion as an unavoidable source of failure and focus on correctly identifying objects upon reappearance \cite{betke, vision_hist}. Alternatively, they utilize features that distinguish each object and rely on warning signs to detect occlusion in advance \cite{vision_slow}. Occupancy grids are a class of multi-object tracking algorithms that forego representation of distinct objects and instead model a region of space \cite{grid_detailed}. A grid of adequate resolution is usually more computationally expensive than a similar multi-object tracker, but grids have the advantage of easily incorporating occlusion and other interaction effects. Recent research has applied theory from object tracking to grids \cite{grid_rfs} and learned grid trackers with techniques from computer vision \cite{grid_deep}. Finally, occlusion has been incorporated into the framework of set-theoretic multi-object tracking. Prior work has focused on one representation of occlusion and run into limitations, typically resorting to handmade approximations. Section \ref{framework} covers the framework of multi-object tracking, and section \ref{occlusionsection} discusses ways to incorporate occlusion into this framework, with the final sections providing two use cases. But first we differentiate approaches to modeling occlusion with a simple example.

\section{Four Square Example} \label{foursquare}
\begin{figure*}[ht]
\centering
\includegraphics[width=6.0in]{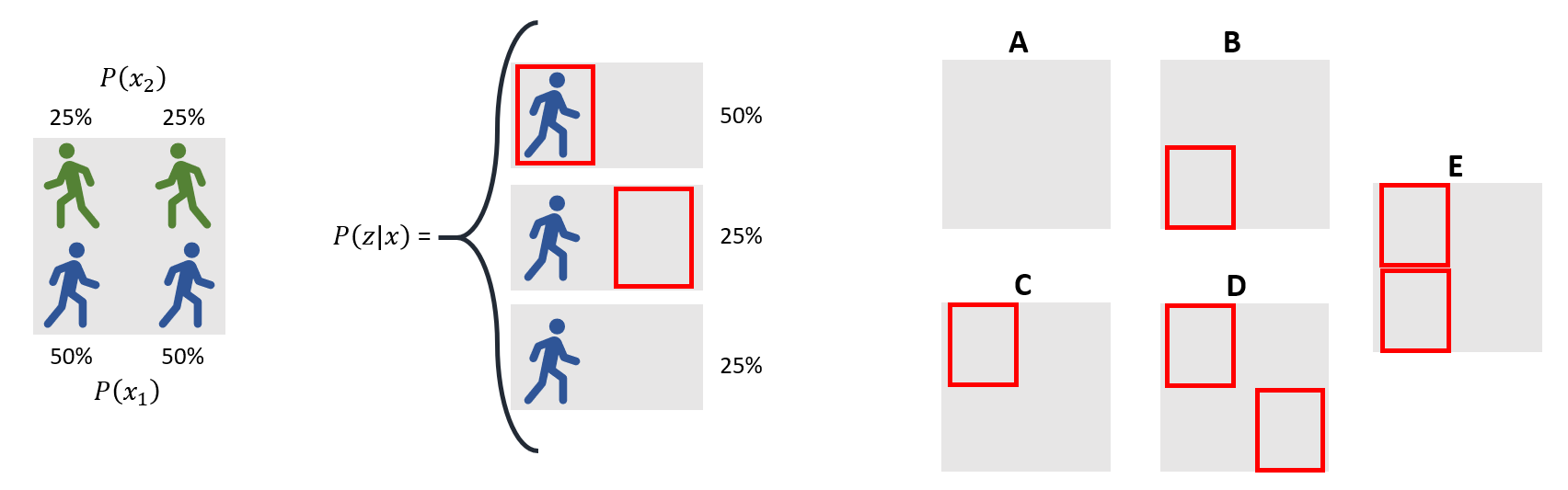}
\caption{Depiction of four-square example: object probabilities (left), measurement model (center), and five example measurement outcomes (right).}
\label{foursquarefig}
\end{figure*}

This example uses a discrete space with up to two objects and measurements. As shown in Figure \ref{foursquarefig}, one object is guaranteed to be present and has an equal chance to exist in either the bottom left or bottom right square. The other object has an equal chance of being present or not present, and if present it has an equal chance of being in the top right or the top left square. A present object has a 50\% chance of generating a measurement in the same square, a 25\% chance of generating a measurement in the wrong square due to hypothetical sensor error, and a 25\% chance of not generating a measurement due to sensor failure. There are no false positives in this example, i.e. a row without an object will not have any measurements. Figure \ref{foursquarefig} displays this model, with measurements denoted as red boxes. Because each object and measurement are consigned to separate rows, if there is no occlusion then the prior, measurement, and posterior distributions can be handled separately for each object. Several possible measurement outcomes are shown in Figure \ref{foursquarefig}, and the posterior estimate of the objects given each outcome is shown in the ``No Occlusion'' rows of Table \ref{foursquareresults}. \par

We next assume that the object in the bottom row may occlude the top one. This example displays a common motivation for tracking under occlusion: to determine the presence and rough location of objects behind currently tracked objects. We first follow the traditional representation of occlusion: if the top object is behind the bottom object, it cannot generate any measurement. This naturally leads to a different posterior estimate, not only for the top object's existence but also for the expected positions of both objects. For instance, the probability of the bottom object being in the left square given outcome D is much lower, because an object in the bottom left square would occlude the object creating a measurement in the top left square.

\begin{figure}[ht]
\centering
\includegraphics[width=2.in]{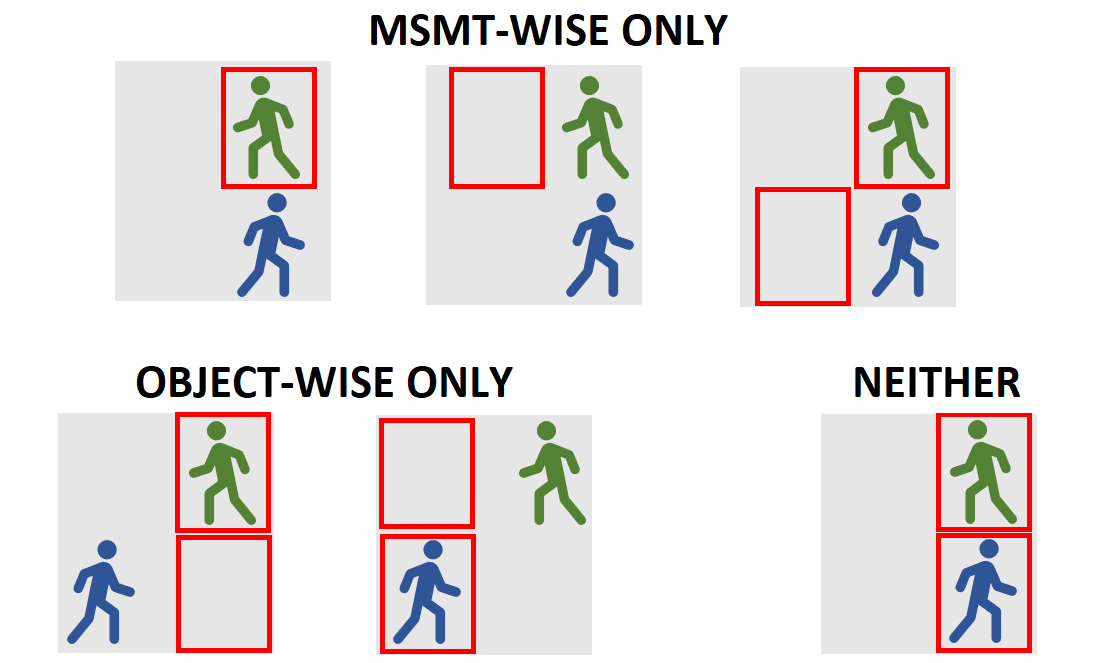}
\caption{Measurement outcomes that are considered possible by one occlusion model but not the other.}
\label{foursquareocclusion}
\end{figure}

In the second representation of occlusion, the placement of objects is irrelevant, but a measurement in the bottom row renders a top measurement in the same column invisible. We refer to the first representation as object-wise occlusion, and the second as measurement-wise occlusion. Despite having similar base concept, they can ultimately have distinct effects on the posterior estimation of either object. Figure \ref{foursquareocclusion} lists the outcomes of this example that are considered impossible by either representation. Table \ref{foursquareresults} includes results from both types of occlusion, which can cause significantly different conclusions. Note that a posterior cannot be derived for measurement set E with measurement-wise occlusion, because such a measurement set is considered impossible. \par

\hfill

\begin{figure}[ht]
\centering
\includegraphics[width=2.5in]{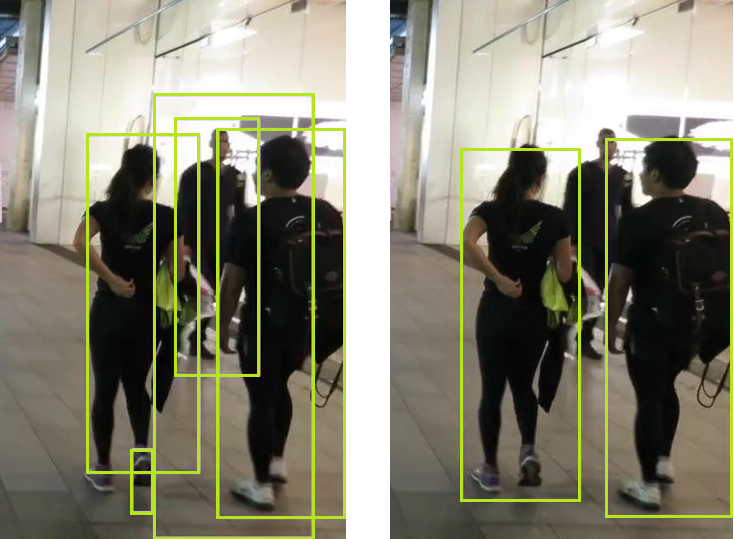}
\caption{Provided detections from the MOT17 benchmark video 10 \cite{mot}. The left detections are from DPM, and includes the partially occluded pedestrian but also includes several false positives. The right detections are from Faster-RCNN, which has high precision but fails to detect the occluded pedestrian.}
\label{twovisions}
\end{figure}

\begin{table}[t]
\centering
\renewcommand{\arraystretch}{1.7}
\caption{Posterior distribution of objects\newline given measurements from figure \ref{foursquarefig}}
\label{foursquareresults}
\begin{tabular}{|l|ccccc|}
\hline & \multicolumn{5}{c|}{P(top object exists)} \\
 & A & B & C & D & E \\
\hline No Occlusion & $\frac{1}{5}$ & $\frac{1}{5}$ & 1 & 1 & 1 \\
Object-wise Occlusion & $\frac{5}{13}$ & $\frac{5}{13}$ & 1 & 1 & 1 \\
Measurement-wise Occlusion & $\frac{1}{5}$ & $\frac{5}{13}$ & 1 & 1 & - \\
\hline & \multicolumn{5}{c|}{P(top object on left if exists)} \\ 
\hline No Occlusion & $\frac{1}{2}$ & $\frac{1}{2}$ & $\frac{2}{3}$ & $\frac{2}{3}$ & $\frac{2}{3}$ \\
Object-wise Occlusion & $\frac{1}{2}$ & $\frac{3}{5}$ & $\frac{2}{3}$ & $\frac{4}{5}$ & $\frac{1}{2}$ \\
Measurement-wise Occlusion & $\frac{1}{2}$ & $\frac{3}{5}$ & $\frac{2}{3}$ & $\frac{2}{3}$ & - \\
\hline & \multicolumn{5}{c|}{P(bottom object on left)} \\
\hline No Occlusion & $\frac{1}{2}$ & $\frac{2}{3}$ & $\frac{1}{2}$ & $\frac{1}{3}$ & $\frac{2}{3}$ \\
Object-wise Occlusion & $\frac{1}{2}$ & $\frac{2}{3}$ & $\frac{1}{3}$ & $\frac{1}{5}$ & $\frac{1}{2}$ \\
Measurement-wise Occlusion & $\frac{1}{2}$ & $\frac{2}{3}$ & $\frac{1}{2}$ & $\frac{1}{3}$ & - \\ \hline
\end{tabular}
\end{table}
Which representation is more valid? For a highly accurate sensor, the outcomes for which object-wise and measurement-wise occlusion differ would rarely occur and the difference becomes trivial. Sensors for which the object-wise representation is better suited include:
\begin{itemize}
\item Sensors that generate a small number of point measurements per object, such as post-processed radar. Even if clustering is used to match one measurement group per object, the definition of a measurement-wise occlusion would be complex and case-specific. Radar, however, requires a complex formulation of occlusion in the first place due to its reflective tendency \cite{thrun,variational_radar}.
\item Computer vision algorithms that can infer the overall position of an object based on individual parts, especially when the occluding objects are nonconvex shapes such as humans. Deformable part-based models are an example. Figure \ref{twovisions} shows an example image where a moderately overlapping person was detected distinctly. Once again, the nature of occlusion for this type of sensor is quite complex.
\end{itemize}
Sensors for which the measurement-wise representation may be more valid include
\begin{itemize}
\item Sensors that give unprocessed, high-resolution information, such as scanning lasers (lidar). These sensors give a fixed number of measurements at known angles, so any hypothetical measurement can only be occluded by a measurement at the same angle. The value of measurement-wise occlusion is especially clear for sensors whose sight is not parallel to the plane in which the objects move, such as rotational lidars placed on drones or the tops of vehicles \cite{lidar3d}. The probability of occlusion for each laser will depend on the height of each object, as well as any elevation or sensor tilt, whereas measurement-wise occlusion can be reasoned about with only a measured range value. We show in section \ref{seplikstuff} that some object-wise approximations for lidar tracking can be handled directly with measurement-wise occlusion.
\item Computer vision algorithms that utilize non-maximum suppression (NMS). Many computer vision techniques give multiple small or overlapping detection responses for a single object. NMS removes or merges overlapping detections to address this problem, at the cost of potentially removing detections of different, nearby objects. In other words, it is an intentional implementation of measurement-wise occlusion. Occlusion-sensitive versions of NMS have been studied \cite{nms}, but to our knowledge have not been heavily adopted. The right side of Figure \ref{twovisions} shows detections from a deep-learning vision algorithm that has utilized NMS.
\end{itemize}

Ultimately, either approach is a simplification of the complex or possibly unknown true behavior of a sensor. The next sections show how these occlusion methods can be implemented for multi-object tracking.

\section{Tracking Framework} \label{framework}
This section briefly describes multi-object tracking, omitting steps that are unaffected by occlusion such as prediction and object creation and removal. $X=\{x_1,...,x_n\}$ is a set of objects $x_i$ that is distributed according to a set probability density function $P(X)$. Similarly, $Z=\{z_1,...,z_m\}$ is a set of measurements $z_j$, generated from $X$ by the likelihood function $P(Z|X)$. The goal of multi-object tracking is to determine, or approximate, the posterior distribution $P(X|Z)$. This parallels the goal of single-object tracking to determine:
\begin{align}
p(x_i|z_j) = \frac{p(z_j|x_i) p(x_i)}{\int_{x_i} p(z_j|x_i) p(x_i)}
\end{align}
and in fact multi-object models are designed to utilize similar pairwise object-measurement relationships. We adopt the disjoint union notation of \cite{pmbm}, in which the probability of a finite, unordered set can be written as a sum of permutations across a fixed-size, ordered list of disjoint subsets.
\begin{equation}
P(X) = \sum_{X_1\uplus X_2 \uplus ... \uplus X_n = X} P(X_1, X_2, ..., X_n)
\end{equation}
The notation $X_1 \uplus X_2 = X$ means that $X_1 \cup X_2 = X$, $X_1 \cap X_2 = \oslash$. This notation has not been widely adopted but offers several conveniences. For instance, probabilities over the superposition of two sets can be cleanly written.
\begin{align}
&Z = X \uplus Y \nonumber \\
&P(Z) = \sum_{X_1 \uplus ... \uplus X_n = X} \sum_{Y_1 \uplus ... \uplus Y_n  = Y} P(Z_1, ..., Z_n) \\
&Z_i = X_i \uplus Y_i \nonumber
\end{align}
\subsection{Object Models}
The distribution $P(X)$ is chosen based on descriptive power, as well as conjugacy with the measurement likelihood. For instance, the multi-bernoulli distribution \cite{basic_rfs} (and the equivalent classical filter JIPDA) describes a set of $N_i$ potential objects with independent probability of existing $r_i$ and independent state distributions $p_i(x)$.
\begin{align}
P(X) = \text{MB}(X|...,r_i, p_i, ...) &= \sum_{X_1\uplus X_2 \uplus ... \uplus X_{N_i} = X} \prod_{i=1}^{N_i} P(X_i) \label{MB} \\
P(X_i) &= \left\{ \begin{array}{lr} r_i p_i(x) & X_i = {x} \\
1 - r_i & X_i = \oslash \\ 0 & |X_i| > 1 \end{array} \right\} \nonumber
\end{align}
We use the multi-bernoulli distribution as an example for the rest of the paper, on the grounds that other distributions have similar forms and reach similar posterior distributions (in the respects that are relevant to occlusion). For instance, the multi-bernoulli mixture filter uses a mixture of multi-bernoulli distributions, the labeled MB and GLMB filter have similar forms, and all of the above can be combined with an independent poisson point process to smoothly handle object appearances \cite{pmb,pmbm}.
\subsection{Measurement Model}
Many sensors return a single measurement corresponding to each successfully detected object. This is represented by a single-measurement likelihood $p(z|x)$ and an object-dependent detection probability $P_D(x)$. Additionally, sensors may return false positive measurements, which are typically assumed to be Poisson distributed with a generation rate $\kappa$ and distribution $p_F(z)$. These assumptions are referred to as the standard measurement model and can be fully written as
\begin{align}
P(Z&|X) = P(Z | {x_1,...,x_n}) \nonumber \\
P(Z&|X) = \sum_{Z_1 \uplus ... \uplus Z_n \uplus Z_F=Z} P(Z_F) \prod_{k=1}^n P(Z_k|x_k) \label {standardMM} \\
&P(Z_k | x_k) = \left\{ \begin{array}{lr} P_D(x_k) p(z|x_k) & Z_k = \{z\} \\
1 - P_D(x_k) & Z_k = \oslash \\ 0 & |Z_k| > 1 \end{array} \right\} \label{individualMM} \\
&P(Z_F) = e^{-\kappa} \prod_{z \in Z_f} \kappa p_F(z)
\end{align}
Note that any number of objects may be assigned to the null measurement (undetected), and likewise any number of measurements may be false positives. The joint probability of the multi-bernoulli object model and the standard measurement model can be factored into a convenient form by rearranging the association variables.
\begin{align}
P(Z,X) = &e^{-\kappa} \sum_{\phi} \prod_{j>0,i>0,\phi_{i,j}=1} r_i \int_x p^z_{ij}(x) \prod_{\phi_{i,0}=1} \lambda^0_i \nonumber \\
& \prod_{\phi_{0,j}=1} \kappa p_f(z_j) \, \text{MB}\left( X|...,r^+_i(\phi), p^+_i(\phi), ...\right) \label{posterior} \\
r^+_i(\phi) = &\left\{ \begin{array}{lr} 1 & \phi_{ij}=1,j>1 \\ \frac{r_i \int_x p^0_i(x)}{\lambda^0_i} & \phi_{i0}=1 \end{array} \right\} \nonumber \\
p^+_i(x,\phi) = &\left\{ \begin{array}{lr} \frac{p_{ij}^z(x)}{\int_x p_{ij}^z(x)} & \phi_{ij}=1,j>1 \\ \frac{p_i^0(x)}{\int_x p_i^0(x)} & \phi_{i0}=1 \end{array} \right\} \nonumber \\
p_{ij}^z(x) = &p_i(x) P_D(x)p(z_j|x) \label{hitterm} \\
p_i^0(x) = &p_i(x) \left(1 - P_D(x)\right) \label{missterm}\\
\lambda^0_i = &1 - r_i + r_i \int_x p_i^0(x)
\end{align}
$\phi$ is a matrix-shaped association variable between bernoulli components and measurements.
\begin{align}
\phi_{ij} \in \{0,1\}& \,\forall\, i=0\cdots N_i, \,j=0\cdots m \nonumber \\
 \sum_i \phi_{ij} = 1& \,\forall\, j>0 ,\; \sum_j \theta_{ij} = 1 \,\forall\, i>0 \nonumber
\end{align}
The posterior distribution of $X$ is a mixture of multi-bernoulli distributions. The number of components in the mixture is equal to the number of possible associations, so in practice approximations of this form are used. The marginal distribution of $Z$ is also evident from \eqref{posterior}, and can be thought of as the marginal of a function over associations $\theta$. Calculation or approximation of this marginal probability can be performed in several ways, for instance using graphical techniques \cite{lbp}.

Some sensors, such as scanning lidars or computer vision techniques that collect simple features, instead generate a fixed number of measurements with an arbitrary number detecting any one object. These sensors could be described by applying the standard measurement model to each measurement separately, and assuming that at most one measurement is viewed for any given model. The separable likelihood model \cite{seplik_old,seplik_new} combines this framework with the assumption that objects are easily separable in the measurement space. It can thus consider the measurement-object matchings as predetermined. Other non-standard models parametrize the rate at which an object creates measurements \cite{multimeasurement}. These models are not covered further because, as mentioned before, measurement-wise occlusion is difficult to formulate for such sensors. Intuitively, the standard and separable likelihood models enforce that the set of measurements is a collection of separate pieces of information about individual objects, with uncertainty only in the completeness and association of this information. Certain formulations of occlusion can threaten this assumption.

\section{Occlusion} \label{occlusionsection}
While discussed heavily in the design of practical multi-target trackers, the phenomenon of occlusion has not (to our knowledge) been formally defined for random sets. We start with a random set $X = \{x_1, ... , x_i, ..., x_n\}$ which follows some distribution $P(X)$. Occlusion divides the original set into two disjoint sets: the visible set $X^V$ and the occluded set $X^U$.
At its most general, a probabilistic occlusion model could be written
\begin{align}
P(X^V , X^U | X) = &\mathbf{I}_{\text{dju}} P(v_1, ..., v_n | x_1, ..., x_n) \\
&\mathbf{I}_{\text{dju}} = \left\{ \begin{array}{lr} 1 & X^U \uplus X^V = X \\ 0 & \text{else} \end{array} \right\}
\end{align}
Where values $v$ represent whether or not a particular element was occluded \footnote{We don't strictly define $v$ as a random variable, just as a useful symbol.}. We next define \textit{restricted occlusion}, in which only visible objects impact the occlusion of other objects:

\begin{figure}[t]
\centering
\includegraphics[width=0.5in]{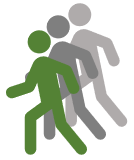}
\caption{Case where the restricted occlusion model is inaccurate. Front object occludes middle object and not back object, but middle object occludes back object.}
\label{restrictedocc}
\end{figure}

\begin{equation}
P(X^V, X^U | X) = \mathbf{I}_{\text{dju}} P \left( V | X^V \right) \prod_{x\in X^U} P(\bar{v} | x, X^V)
\end{equation}
This assumption may not always be realistic: for instance, some sensors may miss objects that are partially occluded even as those objects occlude others, as illustrated in Figure \ref{restrictedocc}. This is however a reasonable assumption in many cases, and is useful for straightforward inference. An even stricter form of occlusion is \textit{static occlusion}:
\begin{equation}
P(X^V, X^U|X) = \mathbf{I}_{\text{dju}} \prod_{x\in X^V} P(v | x) \prod_{x\in X^U} P(\bar{v} | x)
\end{equation}
In this case, no object affects another object's probability of occlusion. This is valid when the causes of occlusion are known rather than being tracked, and is approximately valid when they are tracked very accurately.
\subsection{Object-wise Occlusion}
Object-wise occlusion dictates that from the tracked object set $X$, only a subset of objects $X^V$ are actually capable of generating measurements. Static occlusion in particular can be incorporated into the multi-bernoulli distribution.
\begin{align}
P(X^V, X^U) =& \sum_{X_1 \uplus ... \uplus X_{N_i} = X} \prod_{i=1}^{N_i} \\
&\; \left\{ \begin{array}{lr} r_i p_i(x) p(v|x) & X_i=\{x\},x \in X^V  \\
r_i p_i(x) p(\bar{v}|x) & X_i=\{x\}, x \in X^U \\
1 - r_i & X_i=\oslash \end{array} \right\} \nonumber
\end{align}
The joint probability given the standard measurement model is of the same form as \eqref{posterior}, with the following modifications.
\begin{align}
p^z_{ij}(x) = & p_i(x) p(v|x) P_D(x) p(z_j|x) \\
p^0_i(x) = & p_i(x) p(v|x) \left(1 - P_D(x)\right) + p_i(x) p(\bar{v}|x) \nonumber \\
= & p_i(x) \left(1 - P_D(x)p(v|x) \right) \label{misstermOWO}
\end{align}
It is clear that incorporating static object-wise occlusion in a tracking model is equivalent to modifying the probability of detection to $P_D(x) P(v|x)$. However, the general and even restricted occlusion models are difficult to formulate in such a way: $P(X,V)$ will no longer simply be a product of individual likelihoods for each permutation. \par
Thus trackers use the static occlusion model and alter each object's detection probability, even when the probability of occlusion is highly dependent on nearby objects. The marginal occlusion probability  $p(v|x_i) = \int_{X_{-i}}P(x_i \in X^V)$ where $X_{-i}\uplus \{x_i\} = X$ is the logical choice for a static occlusion term. \cite{negative} accurately solves for the probability of occlusion between two rectangular objects tracked by a line-of-sight sensor, by calculating both the mean and variance of each object's angular span and assuming they are independently distributed. However, this method cannot handle an object that is partially occluded by multiple objects, jointly resulting in a full occlusion. In such situations, they estimate the joint probability of visibility for a given object as the product of these pairwise occlusion probabilities. \cite{occ_integral} handles approximately ellipsoidal objects in a similar way. \cite{expval_softmax} uses the mean position of each object to approximate a static occlusion model, but they calculate the joint probability of occlusion by making a miniature grid across the visible parameter of the rectangle. For a sensor that can handle partial occlusions well, the probability of visibility for the object is the maximum probability of visibility in this grid. \cite{expval_softmax} also uses a exponential weighting to calculate the probability of occlusion for each grid point, to mitigate the inaccuracy of the expected-value approximation. Other practical algorithms such as \cite{stiller} perform deterministic checks for occlusion, assuming that the high accuracy of their sensory data keeps approximation error low.

\subsection{Measurement-wise Occlusion}
Intuitively, measurement-wise occlusion should only affect the probability that an object did not generate one of the visible measurements. Specifically, each object term in the standard measurement model \eqref{individualMM} could be modified to:
\begin{align}
&P(Z_k | x_k, Z^V) = \label{new_individual_MM} \\
&\left\{ \begin{array}{lr} P_D(x_k) p(z|x_k) & Z_k = \{z\} \\
1 - P_D(x_k) + P_D(x_k) \int_z p(z|x_k) p(\bar{v} | z, Z^V) & Z_k = \oslash \end{array} \right\} \nonumber
\end{align}
adding in the probability that object $k$ generates a measurement that was occluded by the visible measurements $Z^V$. This result is in fact obtained under restricted occlusion, regardless of the visibility model $P(V|Z^V)$. The proof uses a convenient property of integration on disjoint sets, proven in \cite{mahler_book} section 3.5.3.
\begin{equation} \label{integral_property}
\int_X \sum_{X_1 \uplus ... \uplus X_n = X} \prod_{k=1}^n f_k(X_k) \,=\, \prod_{k=1}^n \int_{X_k} f_k(X_k)
\end{equation}
The standard measurement model with restricted measurement-wise occlusion can be written: 
\begin{align}
P(Z^V,&Z^U) = P(V|Z^V) \sum_{Z^V_1 \uplus ... \uplus Z^V_n \uplus Z^V_F = Z^V} \\
&\qquad\quad \sum_{Z^U_1 \uplus ... \uplus Z^U_n \uplus Z^U_F = Z^U} \prod_{z \in Z^U_F \cup Z^V_F} \kappa p_F(z) \prod_{k=1}^n \Psi_k \nonumber \\
\Psi_k =& \left\{ \begin{array}{lr} P_D(x_k) p(z|x) & Z_k^V = \{z\},Z_k^U=\oslash \\
P_D(x_k) p(z|x) p(\bar{v}|z,Z^V) & Z_k^V = \oslash,Z_k^U = \{z\} \\
1 - P_D(x_k) & Z_k^V = \oslash , Z_k^U=\oslash \\
0 & \text{else} \end{array} \right\} \nonumber
\end{align}
We are only interested in the probability of the observed measurements $Z^V$, and so integrate $Z^U$ out of each term.
\begin{align}
\int_{Z^U_k}& \Psi_k = P(Z_k | x_k, Z^V) \;\; \text{from (\ref{new_individual_MM})} \nonumber \\
\int_{Z^U_F}& \prod_{z \in Z^U_F \cup Z^V_F} \kappa p_F(z) = e^{\kappa \int_{z} p_F(z) P( \bar{v} | z, Z^V)} \prod_{z \in Z^V_F} \kappa p_F(z) \nonumber
\end{align}
Functionally, the only change to the multi-bernoulli joint distribution is an addition to term \eqref{missterm}.
\begin{equation}
p_i^0(x) = p_i(x) \left(1 - P_D(x) + P_D(x) \int_z p(z|x) p(\bar{v} | z, Z^V) \right)
\end{equation}
In addition to this change, the measurement model is multiplied by a constant exponential term corresponding to occluded false positives, and by $P(V | Z^V)$. In restricted object-wise occlusion, $P(V | X^V)$ would complicate inference by adding inter-object dependencies. In measurement-wise occlusion, the visible measurements are known and so this term is irrelevant to calculation of the posterior.

\section{Separable Likelihood Application} \label{seplikstuff}
Section \ref{foursquare} argued that measurement-wise occlusion is a realistic choice for scanning line-of-sight sensors. Here the potential simplicity of its application is demonstrated. For these sensors, the standard measurement likelihood for each measurement can be written separately:
\begin{align}
p(z|X) =& \sum_{k} c_k(X) p(z|x_k) + c_F(X) p_f(z) \\
c_k(X) =& P_D(x_k) \prod_{k'\neq k} \left( 1-P_D(x_{k'})\right) \nonumber \\
c_F(X) =& \kappa \prod_{k} \left( 1-P_D(x_{k})\right) \nonumber
\end{align}

\begin{figure}[ht]
\centering
\includegraphics[width=2.5in]{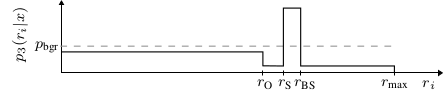}
\caption{Figure taken directly from \cite{seplik_new}, showing part of their occlusion-tolerant measurement model.}
\label{seplikfig}
\end{figure}

This method has been utilized by \cite{thrun, seplik_new, seplik_newest} to track vehicles using horizontally scanning lidar. Each work designed a measurement likelihood that was resistant to occlusion between well-separated objects. As shown in Figure \ref{seplikfig}, measurements near the hypothesized vehicle were highly likely, measurements slightly farther away were highly unlikely, and measurements significantly closer to the sensor were given a moderate, uniform likelihood. Alternatively, consider a deterministic restricted measurement-wise occlusion model where any measurement occludes all measurements with a higher distance. If objects are separated in distance enough that any given measurement is much more likely to have been generated from one object (or be a false positive) than the others, then the multi-bernoulli separable-measurement joint distribution can be simplified greatly.
\begin{align}
p(z,X) \propto\,& \text{MB}\left( X|...,r_i^+(z), p_i^+(z),...\right) \\
r_i^+(z) =& \left\{ \begin{array}{lr} 1 & \hat{z} \approx z \\ \frac{r_i \int_x p_i^0(x)}{\lambda_i^0} & \hat{z} < z \\
r_i & \hat{z} > z \end{array} \right\} \nonumber \\
p_i^+(x|z) =& \left\{ \begin{array}{lr} \frac{p_i^z(x)}{\int_x p_i^z(x)} & \hat{z} \approx z \\ \frac{p_i^0(x)}{\int_x p_i^0(x)} & \hat{z} < z \\
p_i(x) & \hat{z} > z \end{array} \right\} \nonumber \\
&\hat{z} = \frac{ \int_x p_i^z(x) }{1 - \int_x p_i^0(x)} \nonumber
\end{align}
Where $p_i^z(x)$ and $p_i^0(x)$ were defined in \eqref{hitterm} and \eqref{missterm}. Measurement-wise occlusion gives the properties desired by \cite{thrun, seplik_new, seplik_newest}, without their constraints on the measurement likelihood. This permits, for instance, separable-likelihood tracking using Kalman or Rao-Blackwellized filters. Relaxing some of the assumptions, such as separable false positives or the deterministic nature of the occlusion, will still result in a tractable multi-bernoulli mixture posterior, though not necessarily a singular multi-bernoulli.

\section{Visual Tracking Application}

\begin{figure}[ht]
\centering
\subfloat[Raw Detections]{\includegraphics[width=3.4in]{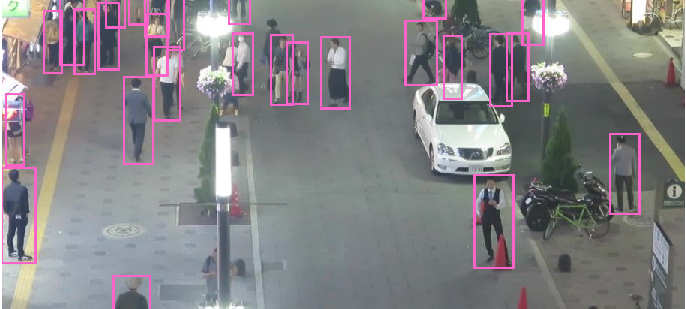}%
}
\\ 
\subfloat[Tracking Estimates (measurement-wise occlusion)]{\includegraphics[width=3.4in]{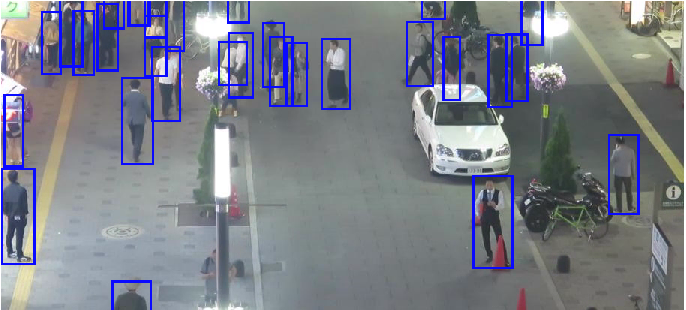}%
}
\caption{A snapshot from the MOT17 Benchmark video 4.}
\label{motsnapshot}
\end{figure}


\begin{table*}[ht]
\centering
\renewcommand{\arraystretch}{1.3}
\caption{Performance metrics of PMB trackers on MOT17 video 4}
\label{motresults}
\begin{tabular}{|l|c|c|c|c|c|c|c|c||c|c|}
\hline Occlusion & MOTA $\uparrow$ & MOTP $\downarrow$ & IDF1 $\uparrow$ & Mostly Tracked $\uparrow$ & Mostly Lost $\downarrow$ & FP $\downarrow$ & FN $\downarrow$ & \# Switches $\downarrow$ & GOSPA $\downarrow$ & Cardinality \\
\hline None & .351 & .408 & .755 & 86 & 0 & 18721 & 12 & 8 & 14002 & -.39 \\
\hline MWO & .426 & .418 & .777 & 78 & 0 & 17339 & 0 & 13 & 13947 & -.36 \\
\hline OWO & .427 & .4 & .777 & 76 & 0 & 17317 & 0 & 9 & 13826 & -.36 \\
\hline \end{tabular}
\end{table*}

To demonstrate the value of occlusion-aware tracking beyond simple LOS sensors, we track pedestrians in the fourth video from the 2017 Multi-Object Tracking Benchmark using the supplied bounding boxes from the Faster-RCNN detector \cite{mot}. These detections have a very low false positive rate but can miss partially occluded people, possibly due to heavy non-maximum suppression. This video is a challenging test of occlusion reasoning. There are many cases of pedestrians occluded by single other pedestrians, groups of other pedestrians, and also street lights and other stationary objects whose existences is not known by the tracker.

The bounding boxes of each person are tracked in image space, in which horizontal and vertical location and size are the features. Occlusion is likely if the overlap between boxes, for instance measured by the intersection area over total area, is high. This representation provides no natural ordering of occlusion for objects, unlike in a ground-plane setting where the relative distance to the sensor distinguishes occluding and occluded objects/measurements. We use two techniques to determine order of occlusion: first we assume that measurement boxes can only be occluded by measurement boxes whose bottom is lower than theirs. For right-side-up cameras detecting grounded objects, this emulates a distance-based ordering. To promote stability in the order of occlusion, each object is given a fifth feature, occludability. An object with a 95\% occludability has a 95\% chance of generating an occludable measurement, which may or may not actually be occluded by another measurement, and a 5\% chance of generating a measurement which cannot be occluded no matter where it is. Given that only occludable measurements can be occluded, the posterior occludability inherently increases for undetected objects and is unchanged for detected objects. In the prediction step, occludability is slowly mixed to its equilibrium value. This approach to occlusion is applied to a measurement-wise tracker and to an object-wise tracker using an expected-value approximation. The same tracker is also run without occlusion reasoning.

The object state (sans occludability) is normally distributed, with single-object tracking carried out by a standard Kalman filter. The poisson multi-bernoulli filter \cite{pmb} was used as the multi-object framework, with merging by track so that object labels were kept consistent. The data association step was achieved with the loopy belief propagation technique from \cite{lbp}. For implementation, a fixed array of 2048 normal components and an array of 72 object labels was used. The most likely 2048 components from each update step were kept. Likewise, the most likely 72 objects were kept while the others were `recycled' as unlabeled, poisson-distributed components. Highly similar components in the same object were located via kd-trees and trivially merged by pooling their existence probability. New pedestrians entering the scene are assumed to be poisson-generated at the edges of the image.\par
Table \ref{motresults} shows the accuracy and precision scores used by the MOT benchmark for labeled tracking evaluation, as well as the generalized optimal subpattern assignment metric (GOSPA) \cite{gospa} and ratio of difference in total cardinality as unlabeled performance indicators. Arrows by each metric name indicate the direction of higher performance. Both labeled and unlabeled multi-object metrics require a base single-object metric: bounding box intersection-over-union was chosen as in the MOT15-17 benchmarks, but with a looser cutoff such that any degree of intersection is considered a possible match. The bounding boxes in video 4 are smaller than most in MOT17, and occluded individuals moving in crowded areas would be extremely difficult to match with the standard requirement of 0.5 IoU. As the primary application of the MOT benchmark is consistent post-processed labeling, its standard scoring code removes a significant number of individuals that are heavily occluded or unmoving at each time. We include all of these individuals as our goal is to track temporarily occluded objects.

\begin{figure}[t]
\centering
\includegraphics[width=3.in]{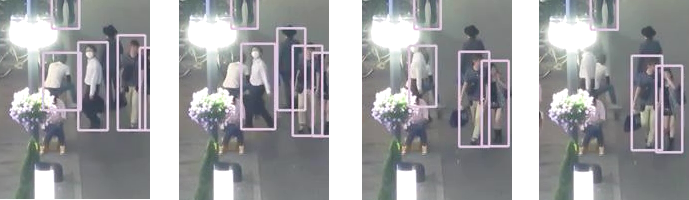}
\caption{Raw detections from MOT17 video 4, increasing in time from left to right. The leftmost person is occluded by someone who also becomes occluded.}
\label{restrictedreal}
\end{figure}

While no tracker has excellent results, the occlusion-equipped models outperform the baseline model by most metrics. The two snapshots of the video in Figure \ref{motsnapshot} show the raw F-RCNN detections in magenta and the hypothesized objects in blue. The crowd in the upper left is not resolved (some individuals here are not detected throughout the video), but the two occlusion cases in the center are easily resolved based on the past positions of these individuals. The approximate object-wise tracker outperforms the measurement-wise tracker, especially in identity switches. It is possible that violation of the restricted occlusion assumption, by the undetected stationary obstacles, significantly impacts measurement-wise tracker. Figure \ref{restrictedreal} shows a case where one person is occluded by another, who proceeds to be occluded by a light pole.

\section{Conclusion}
The traditional formulation of occlusion in multi-object tracking is that objects block other objects from the sensor's view, and that occluded objects generate no measurement. This is intuitive but creates object dependencies that make tracking intractable, so a variety of approximations have been proposed. We instead formally define occlusion as an operation on a random set and show that this operation can be applied to measurements as well as objects. This new approach, termed measurement-wise occlusion, is equally intuitive and fits tractably into the standard multi-object model with a loose restriction. It can be implemented with a simple additional step in any given multi-object tracking technique. We highlighted the practical value of this approach in two tracking applications where occlusion is a significant problem.


\section*{Acknowledgment}
This work was supported by the Texas Department of Transportation under Project 0-6877 entitled ``Communications and Radar-Supported Transportation Operations and Planning (CAR-STOP).''




\bibliographystyle{IEEEtran}
\bibliography{IEEEabrv,main}

\appendix

\subsection{Highway simulations}
\begin{figure}[t]
\centering
\subfloat[Object-wise Simulation]{\includegraphics[width=3.4in]{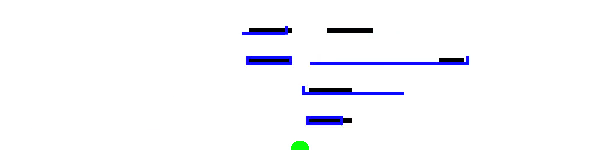}%
}
\\ 
\subfloat[Measurement-wise Simulation]{\includegraphics[width=3.4in]{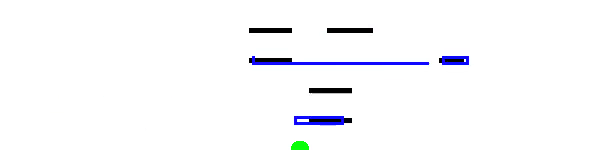}%
}
\caption{A snapshot of the highway simulation. The black lines represent vehicles, while the green circle shows the sensor's position. The blue squares show complete measurements, while the blue lines show measurements that were partially occluded. The lines span the occluded area in which the partially detected vehicle could extend.}
\label{frogger_snap}
\end{figure}

We also create a simple simulated highway to assess occlusion handling for tracking vehicles across multiple lanes\footnote{This section is not in the published version of this paper.}. The highway has four lanes, and in each lane vehicles move at a constant velocity on the center line, much like in the classic arcade game Frogger. A point sensor at the side of the highway views these vehicles. The vehicles' widths are neglected, so their visibility depends entirely on their lane and relative angle from the sensor. For example, say there is a vehicle in the lane nearest the sensor with its back end directly in front of the sensor, and its front end at an angle $\theta$ ahead of the sensor. Under object-wise occlusion, any vehicles in further lanes whose front and back ends lie within $0$ and $\theta$ will be completely occluded. The sensor is assumed to recognize contiguous shapes, so measurement-wise occlusion operates similarly. Missed detections, false positives, and gaussian noise are applied to the sensor output in addition to occlusion. Figure \ref{frogger_snap} visualizes a single timestep of this highway, with two possible random measurement sets corresponding to the two occlusion types.

A particle filter version of the track-oriented multi-bernoulli filter is used so that closed-form updates can be performed even for partially occluded measurements. Measurement-wise occlusion probabilities can also be determined exactly, while object-wise occlusion is approximated in two different ways. The first takes the expected value of potentially occluding objects and calculates the probability that each individually occludes the target object, then combines the individual probabilities with the softmax function as in \cite{expval_softmax}. The second stores a grid representation of the sensor's field of view, and updates the visibility of each cell in the grid based on vehicle positions. Simulation parameters such as the magnitude of measurement noise are known to the tracker. The tracker is run for 10000 timesteps, representing over half an hour of traffic at 5 timesteps per second.

Table \ref{froggerresults} shows performance of each occlusion model in terms of average GOSPA per timestep. Euclidean distance in position and length is used as the base metric. The approximate object-wise occlusion tracker work equally well under either simulated from of occlusion, with the grid approximation outperforming the expected-value approximation. The measurement-wise occlusion tracker scores slightly lower (better) than the grid approximation when the simulated occlusion type matches its assumptions, and slightly higher when object-wise occlusion is simulated. It is worth noting that this simulation is simple enough that an accurate grid approximation can be applied in real time, while more complex applications may not be able to apply it as quickly. Expected-value approximations are fast, but perform worse than the measurement-wise tracker for both simulations. Codes for the simulated tests and for the pedestrian tracking tests are available at \texttt{https://github.com/utexas-ghosh-group/carstop/tree/master/MWO}.

\begin{table}[t]
\centering
\renewcommand{\arraystretch}{1.3}
\caption{Performance of MB trackers on highway simulation.}
\label{froggerresults}
\begin{tabular}{|l|l||c|c|}
\hline Simulated Occlusion & Tracker & GOSPA $\downarrow$ \\
\hline OWO & OWO-expval & 4.94 \\
\hline OWO & OWO-grid & 4.79 \\
\hline OWO & MWO & 4.90 \\
\hline \hline MWO & OWO-expval & 4.94 \\
\hline MWO & OWO-grid & 4.80 \\
\hline MWO & MWO & 4.67 \\
\hline \end{tabular}
\end{table}


\end{document}